\documentclass[10pt,twocolumn,letterpaper]{article}

\usepackage{cvpr}              
\definecolor{cvprblue}{rgb}{0.21,0.49,0.74}
\usepackage[pagebackref,breaklinks,colorlinks,allcolors=cvprblue]{hyperref}
\def\paperID{  } 
\def\confName{CVPR}
\def\confYear{2026}
\usepackage{graphicx}
\usepackage{amsmath}
\usepackage{amssymb}
\usepackage{booktabs}
\usepackage{tcolorbox}
\usepackage[table]{xcolor} 
\definecolor{lightgray}{gray}{0.92}
\usepackage{multirow}  
\usepackage{enumitem}
\usepackage[pagebackref,breaklinks,colorlinks]{hyperref}
\usepackage[ruled,vlined]{algorithm2e}
\usepackage{xcolor}
\usepackage{tabularx}
\usepackage{array}
\usepackage{stfloats}
\usepackage[capitalize]{cleveref}
\crefname{section}{Sec.}{Secs.}
\Crefname{section}{Section}{Sections}
\Crefname{table}{Table}{Tables}
\crefname{table}{Tab.}{Tabs.}
\usepackage{listings}
\usepackage[utf8]{inputenc}
\tcbuselibrary{skins, breakable}
\definecolor{boxframe}{RGB}{150,150,150} 
\definecolor{boxback}{RGB}{255,255,255}  
\definecolor{titlebg}{RGB}{220,220,220}  
\definecolor{jsonkey}{RGB}{0,0,128}      
\definecolor{jsonstring}{RGB}{0,100,0}   

\newtcolorbox{innerbox}{
    enhanced,
    colback=boxback,
    colframe=boxframe,
    boxrule=1pt,
    arc=4pt,             
    boxsep=2pt,
    left=4pt, right=4pt, top=4pt, bottom=4pt,
    fonttitle=\bfseries\small,
    nobeforeafter,       
    bottom=5pt
}

\newtcolorbox{outerbox}[1][]{
    enhanced,
    colframe=black,      
    colback=white,
    coltitle=black,
    boxrule=1.5pt,
    fonttitle=\bfseries\large,
    attach boxed title to top left={yshift=-2mm, xshift=2mm}, 
    title={#1},
    sharp corners,       
    top=5pt, bottom=5pt, left=5pt, right=5pt,
    frame style={draw=black, line width=1.5pt},
    title style={fill=titlebg, draw=black, line width=0pt},
    overlay={
        \draw[black, line width=1pt] (frame.north west) -- (frame.north east);
    }
}
\newcommand{\role}[1]{\noindent\textbf{#1: }}

\title{IntroSVG: Learning from Rendering Feedback for Text-to-SVG Generation via an Introspective Generator–Critic Framework}

\author{
Feiyu Wang$^{1,2*}$ \quad
Jiayuan Yang$^{3}$ \quad
Zhiyuan Zhao$^{2\dagger}$ \quad
Da Zhang$^{2,3}$ \\
Bingyu Li$^{2,4}$ \quad
Peng Liu$^{1}$ \quad
Junyu Gao$^{2,3}$\\
$^{1}$Fudan University \quad
$^{2}$TeleAI \\
$^{3}$Northwestern Polytechnical University \quad
$^{4}$University of Science and Technology of China\\
\href{https://gitcat-404.github.io/IntroSVGProject/}{https://gitcat-404.github.io/IntroSVGProject/}
}

\begin{document}

\twocolumn[{%
\renewcommand\twocolumn[1][]{#1}%
\maketitle
\thispagestyle{empty}

\begin{center}
    \centering
    \includegraphics[width=\textwidth]{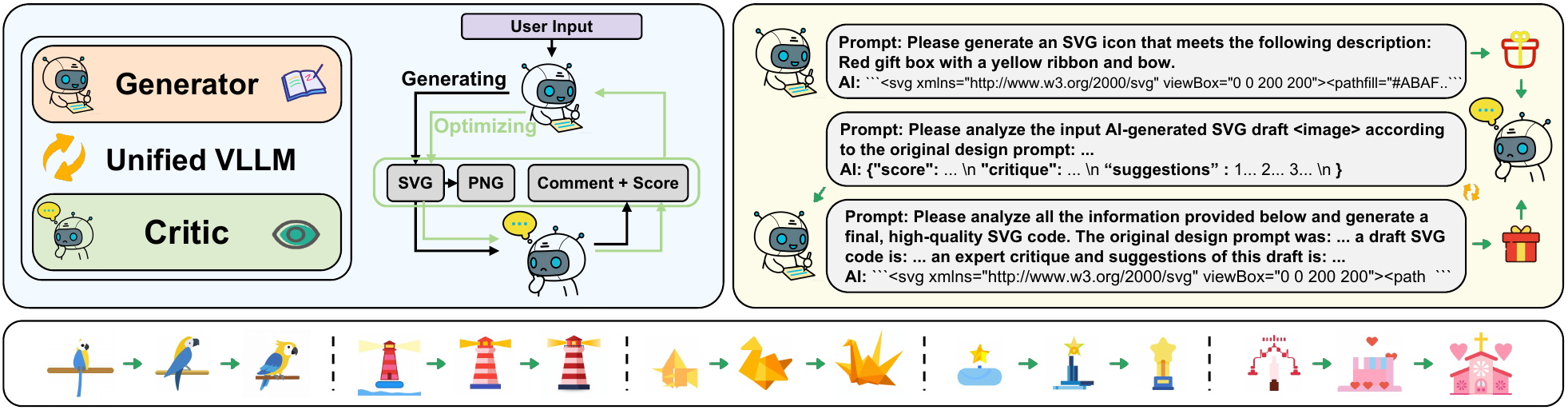}
    \captionof{figure}{Overview of our proposed IntroSVG (Introspective SVG Generation) framework. \textbf{(Left)} At the core is a unified VLM that fulfills the dual roles of "Generator" (drafting SVG) and "Critic" (perceiving PNG feedback). Black arrows represent the initial generation, while green arrows denote the iterative optimization. \textbf{(Right)} The "generate-critique-refine" iterative loop is shown: the model generates an initial draft, self-critiques the rendered PNG, and finally revises the code based on the structured feedback. \textbf{(Bottom)} Visualizations demonstrate how the model autonomously improves a sketch into a high-quality SVG through iterative refinement.}
    \label{fig:teaser}
\end{center}%
}]

\begingroup
\renewcommand{\thefootnote}{\fnsymbol{footnote}}
\setcounter{footnote}{1}
\footnotetext{Work done during an internship at TeleAI.}
\setcounter{footnote}{2}
\footnotetext{Corresponding author.}
\endgroup
\begin{abstract}
Scalable Vector Graphics (SVG) are central to digital design due to their inherent scalability and editability. Despite significant advancements in content generation enabled by Visual Language Models (VLMs), existing text-to-SVG generation methods are limited by a core challenge: the autoregressive training process does not incorporate visual perception of the final rendered image, which fundamentally constrains generation quality. To address this limitation, we propose an Introspective SVG Generation Framework (IntroSVG). At its core, the framework instantiates a unified VLM that operates in a closed loop, assuming dual roles of both generator and critic. Specifically, through Supervised Fine-Tuning (SFT), the model learns to draft SVGs and to provide feedback on their rendered outputs; moreover, we systematically convert early-stage failures into high-quality error-correction training data, thereby enhancing model robustness. Subsequently, we leverage a high-capacity teacher VLM to construct a preference dataset and further align the generator's policy through Direct Preference Optimization (DPO). During inference, the optimized generator and critic operate collaboratively in an iterative "generate-review-refine" cycle, starting from imperfect intermediate drafts to autonomously improve output quality. Experimental results demonstrate that our method achieves state-of-the-art performance across several key evaluation metrics, generating SVGs with more complex structures, stronger semantic alignment, and greater editability. These results corroborate the effectiveness of incorporating explicit visual feedback into the generation loop.
\end{abstract}

\section{Introduction}
Scalable Vector Graphics (SVG) \cite{quint2003scalable,peng2000scalable} constitute a foundational technology in modern web technologies and professional graphic design owing to their resolution independence and editability. 
Recently, the proliferation of Artificial Intelligence Generated Content (AIGC) is driving research into automated Text-to-SVG (T2S) generation, and existing methods primarily follow two technical paths: optimization-based approaches and autoregressive sequence-based approaches. 

Optimization-based approaches ~\cite{xing2023diffsketcher,jain2023vectorfusion,zhang2024text} generate SVGs by iteratively optimizing differentiable rasterizers. 
However, they often entail high computational costs and yield disorganized SVG code that lacks editability. 
In contrast, autoregressive approaches ~\cite{wu2023iconshop,chen2024svgbuilder,rodriguez2023starvector} employ large language models (LLMs) to directly generate SVG code sequences, which preserves \textit{vector editability} and enhancing \textit{practical usability} and has gradually become the mainstream direction in Text-to-SVG (T2S) research. \cite{wang2025svgen,xing2025reason}

Although existing methods achieve satisfactory progress, our investigation identifies several limitations. 
First, most of them focus only on the performance of SVG code sequence generation and neglect improvements in the model’s SVG visual quality assessment, which in turn leaves the model lacking the \textbf{\underline{eye} for perceiving} structured visual feedback. 
Second, the prevailing \textit{one-pass} generation paradigm lacks effective iterative feedback and rely on subsequent manual selection, which leaves the model lacking the \textbf{\underline{mind} for self-evaluation and iterative refinement}. Recent work has also explored feedback-driven optimization for Image-to-SVG generation (I2S)  \cite{rodriguez2025rendering}, which differs from our text-driven formulation.

To this end, we propose an Introspective Synthesis Framework(\textbf{IntroSVG}), built upon a unified VLM to enhance the model’s capacity to generate, perceive, and iteratively refine SVGs by: 
\begin{enumerate}[leftmargin=*]
\item integrating structured visual feedback into the generation process to make the model have the \textit{eye to perceive};  
\item introducing an internal evaluation-correction mechanism make the model have \textit{mind to self-improve}.  
\end{enumerate}

Our framework follows a two-stage evolutionary process.
In the first stage, we introduce a multi-task paradigm in which our model simultaneously acts as a \textit{Generator} and a \textit{Critic}.
The \textit{Generator} is responsible for two core tasks: directly generating SVG code from text prompts and optimizing existing SVG code based on rendered images, suggestions, and scores.
The \textit{Critic} provides "critical feedback" and "actionable revision suggestions" based on the init requirements and the current rendered SVG.
The \textit{Critic} and \textit{Generator} iteratively optimize in a continuous loop until the result meets expectations.
This process produces a series of "generate-review-refine" triplets that are jointly used during training. By learning from these triplets, the system jointly acquires the abilities to \textbf{generate}, \textbf{evaluate}, and \textbf{correct}, thereby achieving \textbf{self-improvement} in SVG synthesis.

Subsequently, we employ an external expert model to construct a preference dataset and apply Direct Preference Optimization (DPO) \cite{rafailov2023direct} to further align the generator's policy, thereby enabling it to internalize preferences for "excellent design." 
Finally, during inference, the optimized Generator and Critic collaborate to execute an iterative cycle that enhances generation quality. Our primary contributions are as follows:

\begin{itemize}
    \item \textbf{Introspective Synthesis Framework}: We design a unified VLM that simultaneously serves as both a Generator and a Critic. This integration enables the model to perform iterative self-optimization by incorporating explicit visual feedback into the generative loop.
    \item \textbf{Learning-from-errors data and optimization engine}: Rather than discarding suboptimal or failed samples produced by the model, we systematically transform them into high-value training signals. During the SFT phase, they serve as "error-correction" data (Sec.~\ref{sec:generator}); during the DPO phase, they are constructed as critical negative preference pairs (Sec.~\ref{sec:Stage 2}) for policy optimization; and during inference, they form the starting point for iterative refinement (Sec.~\ref{sec:Stage 3}).
    \item \textbf{SOTA performance on comprehensive benchmarks}: Experiments demonstrate that our method significantly outperforms existing models on a unified test set derived from prior SOTA projects (LLM4SVG, OmniSVG, and SVGen) across multiple key metrics. Our approach generates complex SVG artifacts with superior aesthetic quality and semantic alignment.
\end{itemize} 


\section{Related Works}
\subsection{Text-to-SVG Generation}
SVG generation comprises two primary sub-tasks: Text-to-SVG and Image-to-SVG  \cite{li2020differentiable,lopes2019learned,reddy2021im2vec}. Text-to-SVG generation directly converts natural-language descriptions into SVG code. Early methods primarily focus on simple graphics or icons. With the maturation of deep learning, two mainstream technical approaches dominate the field: optimization-based methods and direct generation methods.

Optimization-based methods do not directly generate code; instead, they treat SVG path parameters as optimizable variables. They render the graphics to raster images and evaluate them with models such as CLIP to quantify text–image alignment, guiding parameter updates based on the scores. For instance, ClipDraw \cite{frans2022clipdraw} and Clipasso \cite{vinker2022clipasso} use CLIP \cite{radford2021learning} to optimize vector sketches, while SVGDreamer \cite{xing2024svgdreamer} and VectorFusion \cite{jain2023vectorfusion} leverage prior knowledge from diffusion models \cite{rombach2022high} to guide the optimization process, thus producing high-quality visual results. Chat2SVG \cite{wu2025chat2svg} first employs large language models to generate SVG templates containing basic geometric primitives and then conducts a two-stage optimization guided by image diffusion models. However, these methods are computationally intensive, and the resulting SVG code is disorganized and difficult to edit.

Direct generation methods formulate Text-to-SVG as a sequence-to-sequence translation task, leveraging large (visual) language models to directly generate SVG code. LLM4SVG \cite{xing2024empowering} and StarVector \cite{rodriguez2023starvector} represent early explorations in this direction. These approaches fine-tune large (visual) language models on large-scale datasets and define specialized SVG tokens to help LLMs better capture SVG structure. SVGen \cite{wang2025svgen}, Reason-SVG \cite{xing2025reason}, and SVG-Thinker \cite{chen2025svgthinker} incorporate Chain-of-Thought reasoning to enhance interpretability during generation, enabling models to articulate design steps before emitting code. OmniSVG \cite{yang2025omnisvg} proposes a unified framework that leverages VLM to handle multimodal inputs (text, images, and character references) and achieves complex SVG generation, including anime characters.

\subsection{Model Alignment and Preference Optimization}
To overcome the limitations of purely supervised fine-tuning (SFT), researchers increasingly incorporate reinforcement learning (RL) as a principled framework to enhance the inference and structured decision-making capabilities of LLMs, particularly for tasks that require multi-step reasoning and complex decision-making \cite{ouyang2022training,shinn2023reflexion,lewkowycz2022solving}. Among RL algorithms, Proximal Policy Optimization (PPO) \cite{schulman2017proximal} is widely adopted due to its stability and efficiency. One of its variants, Grouped Relative Policy Optimization (GRPO) \cite{shao2024deepseekmath}, is particularly well suited to tasks such as SVG generation, which can be evaluated using rule-based or heuristic signals. For instance, SVGen \cite{wang2025svgen} and Reason-SVG \cite{xing2025reason} use the GRPO algorithm with custom rewards targeting code integrity, semantic accuracy, and inference processes to correct defects remaining after SFT. Unlike these methods that rely on reward engineering, our framework adopts Direct Preference Optimization (DPO) \cite{rafailov2023direct}. We use DPO to optimize the policy of the 'Generator' role, aiming to significantly enhance its 'first-shot generation' quality. This provides a higher-quality starting point for the subsequent introspective refinement, thereby ensuring the quality of the final output.

\begin{figure*}[h] 
  \centering
  \includegraphics[width=\textwidth]{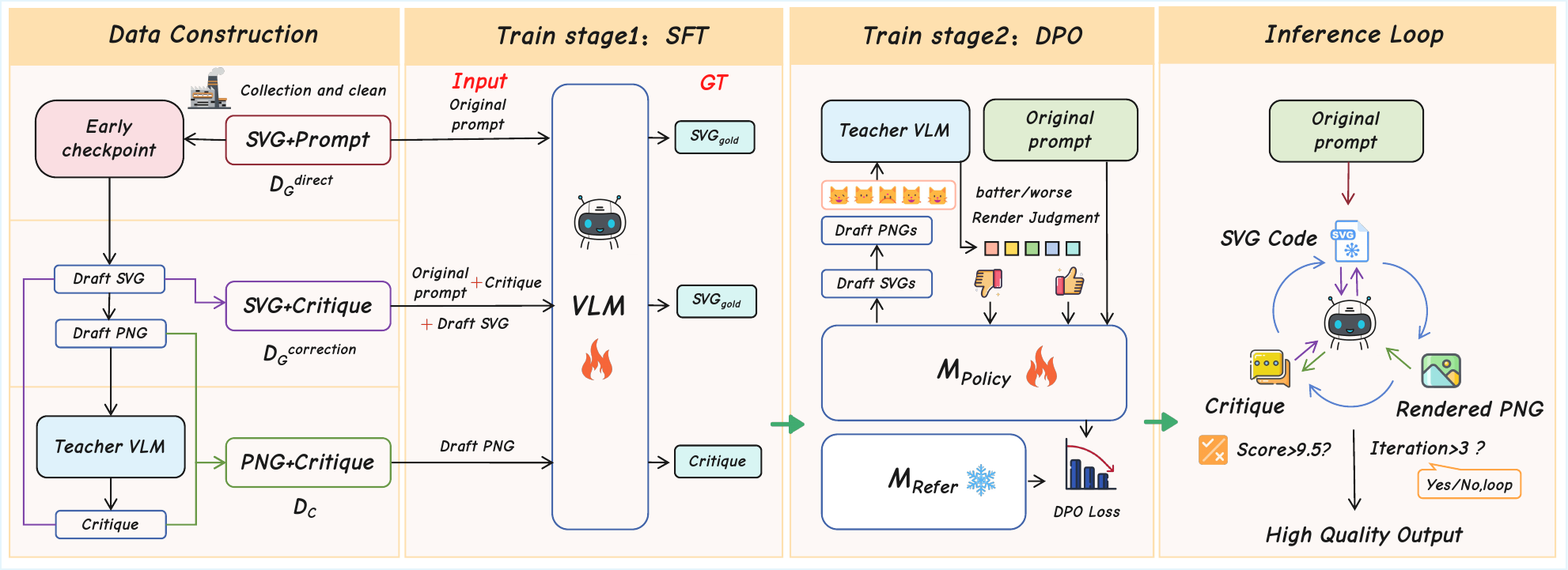} 
  \caption{Overview of the IntroSVG Framework. Our method is divided into the following stages: (\textbf{Data Construction}): Synthesize a mixed dataset for direct generation ($D_G^{\text{direct}}$), correction ($D_G^{\text{correction}}$), and critique ($D_C$) using an early checkpoint model and a Teacher VLM. \textbf{Stage 1 (SFT)}: Train a unified VLM on this mixed dataset, enabling it to possess both generation and critique capabilities simultaneously. \textbf{Stage 2 (DPO)}: Use the Teacher VLM to evaluate generated preference pairs, specifically optimizing the model's generation policy ($M_{\text{Policy}}$) via the DPO loss. \textbf{Introspective inference Loop}: The final single model performs a closed loop during inference: it first generates an SVG, then switches to a Critic role to "view" its rendering and assign a score. If the score is unsatisfactory, it utilizes this critique for the next round of correction.}
  \label{fig:method} 
  \vspace{-0.3cm}
\end{figure*}
\subsection{SVG Datasets and Benchmarks}

To advance the programmatic generation and understanding of vector graphics, several key datasets and benchmarks are now available in the field. Early research typically focuses on datasets tailored to specific tasks. For instance, projects such as FIGR-8-SVG \cite{clouatre2019figr} and DeepSVG \cite{carlier2020deepsvg} develop large-scale monochrome SVG datasets, while SVGBuilder \cite{chen2024svgbuilder} introduces a colorful SVG dataset with hundreds of thousands of samples, thus addressing the gap in color representation. Subsequently, projects like LLM4SVG \cite{xing2024empowering} and StarVector \cite{rodriguez2023starvector} utilize large models to create text pairings for vast numbers of SVGs, though data quality varies. To meet the needs of modern multimodal large language models (MLLMs), recent high-quality, task-unified datasets are developed. These include the SVG-1M dataset with Chain-of-Thought (CoT) \cite{wei2022chain} annotations from the SVGen \cite{wang2025svgen} project, the SVGX-DwT-10k \cite{xing2025reason} dataset which features "Design with Thought" (DwT) annotations from the Reason-SVG project, and the MMSVG-2M \cite{yang2025omnisvg} dataset from the OmniSVG project, which covers illustrations and complex anime characters.

Meanwhile, numerous benchmarks exist to systematically evaluate model capabilities. VGBench \cite{zou2024vgbench} and SVGEditBench-v1/v2 \cite{nishina2024svgeditbench,nishina2025svgeditbench} serve as early exemplars that focus on understanding and editing tasks. Recent benchmarks such as UniSVG \cite{li2025unisvg}, SVGenius \cite{chen2025svgenius}, MMSVG-Bench \cite{yang2025omnisvg}, and ArtifactsBench \cite{zhang2025artifactsbench} not only provide evaluation criteria but also address the limitations of earlier benchmarks by leveraging large-scale training data. 
These benchmarks consistently highlight a core problem: existing models lack the visual feedback and iterative optimization mechanisms inherent to human designers. While they can handle simple edits, their "one-shot" output model struggles to guarantee both visual fidelity and semantic accuracy in complex generation tasks.

This underscores the urgent need for a framework capable of self-inspection and progressive optimization, which directly motivated our research.  We integrated and expanded existing data sources to construct a large-scale, high-quality, multicolor SVG generation dataset and an accompanying correction dataset, providing a solid data foundation for our proposed framework.

\section{Datasets}
\subsection{Data Collection and Cleaning}
Although existing large-scale SVG datasets are abundant, they often exhibit substantial redundancy and high sample similarity, which unnecessarily consume computational resources during training and increase the risk of model overfitting. Furthermore, these datasets exhibit inconsistencies in viewBox dimensions, in coordinate precision (including the number of decimal places), and in the mixed use of relative and absolute path commands. To address these issues, we curate a high-quality, standardized color dataset. It integrates mainstream open-source resources to support complex SVG icon generation while mitigating overfitting risk. We integrate three large-scale open-source datasets from the LLM4SVG \cite{xing2024empowering}, OmniSVG \cite{yang2025omnisvg}, and SVGen \cite{wang2025svgen} projects and employ a rigorous filtering and standardization pipeline: we remove monochrome and non-renderable samples, as well as those with sequence lengths exceeding 8000 tokens; we normalize the viewBox of all samples to “0 0 200 200” and convert basic shapes (e.g., rect, circle) into path elements using absolute path commands; to ensure concise and consistent path representations, we retain only five command types (M, L, C, A, Z) and standardize all coordinates to integers; and we standardize file headers and prefix the fill (fill color) attribute before the d (path data) attribute within each path tag to establish a consistent generation sequence. This process produces approximately 200,000 high-quality (Text prompt, SVG code) pairs.

\subsection{Data Pair Construction}
First, to characterize the model's core generative capabilities, we construct a foundational SFT dataset ($D_G^{\text{direct}}$). This dataset consists of direct pairs of SVG code and their corresponding textual descriptions. Second, to train the model's "correction" and "critique" abilities, we synthesize SFT datasets that target these capabilities. We employ a model pre-trained on $D_G^{\text{direct}}$ to generate SVG drafts for 50,000 prompts. Subsequently, we employ GPT-4o as an external expert to analyze these drafts and their corresponding rendered images, thereby producing JSON feedback containing {score, critique, suggestions}. Based on this feedback, we construct two additional datasets: a dataset of approximately 50,000 critique-training samples ($D_C$), where the input is the original prompt and the rendered image, and the output is the expert's JSON critique; and a dataset of approximately 50,000 correction-training samples ($D_G^{\text{correction}}$) to train the generator's correction ability, where the input is the original prompt, the SVG draft, and the expert's JSON critique, and the output is a high-quality reference SVG from $D_G^{\text{direct}}$. We merge these two datasets with $D_G^{\text{direct}}$ to form the complete training data for the SFT phase: $D_{\text{SFT}} = D_G^{\text{direct}} \cup D_G^{\text{correction}} \cup D_C$. Finally, for Direct Preference Optimization (DPO), we construct a preference dataset ($D_{\text{pref-G}}$). We select 10,000 prompts and use the SFT-tuned model ($M_{\text{SFT}}$, trained on $D_{\text{SFT}}$) to generate 5 distinct SVG candidate samples per prompt (50,000 total). We then employ GPT-4o as an evaluator to score these 50,000 samples. We automatically construct the final preference dataset, $D_{\text{pref-G}}$, by applying two rules: "Render-Success Priority" (a renderable sample is always preferred over a non-renderable one) and "High-Score Priority" (for two renderable samples, the one with the higher expert score is preferred). This process generates preference pairs of the form (prompt, winning sample $S_w$, losing sample $S_l$). Specific data processing details can be found in the "Data Construction" section of the Appendix. 

\section{Method}
A unified Vision-Language Model (VLM) $\mathcal{M}$, parameterized by $\theta$, forms the core of our framework. Through a multistage evolutionary process, $\mathcal{M}$ acquires dual capabilities: generation and critique. Our approach consists of three phases: supervised fine-tuning (SFT) for capability training, direct preference optimization (DPO), and, finally, introspective inference. The overview of our method is depicted in Figure ~\ref{fig:method}.
\subsection{Stage 1: SFT Capability Training}
\label{sec:Stage 1}
The objective of this stage is to instill foundational "Generation" and "Critique" capabilities into the model using the mixed dataset $D_{\text{SFT}}$ defined in the previous section. This is accomplished through two parallel SFT objectives.
\subsubsection{Training "The Generator"}
\label{sec:generator}
We train the generative capability using the $D_G = D_G^{\text{direct}} \cup D_G^{\text{correction}}$ dataset. The objective is to minimize the standard Negative Log-Likelihood (NLL) loss, $\mathcal{L}_{\text{SFT-G}}$:$$\mathcal{L}_{\text{SFT-G}}(\theta) = - \mathbb{E}_{(X_G, S_{gold}) \sim D_G} [\log p(S_{gold} | X_G; \theta)]$$where $S_{gold}$ is the high-quality reference SVG. Crucially, the input $X_G$ has two forms: it can be a simple prompt $P$ from $D_G^{\text{direct}}$, or a complex correction prompt $P_{\text{complex}}$ (containing $P$, $S_{\text{fail}}$, $C_{\text{fail}}$) from $D_G^{\text{correction}}$. This allows the model $\mathcal{M}$ to not only learn creation from scratch but also to internalize the ability to "correct from mistakes".
\subsubsection{Training "The Critic"}
\label{sec:critic}
We use the critique dataset $D_C$ to train the model to evaluate outputs and provide feedback. As described in the "Datasets" section, this dataset contains images $I$ that are rendered from $S_{\text{fail}}$. The training objective is to minimize the NLL loss $\mathcal{L}_{\text{SFT-C}}$, enabling the model to predict the expert's structured critique $C$:
$$\mathcal{L}_{\text{SFT-C}}(\theta) = - \mathbb{E}_{(P, I, C) \sim D_C} [\log p(C | P, I; \theta)]$$
Through this stage, the model $\mathcal{M}_{\text{SFT}}$ learns to output "aesthetic judgments" and "revision suggestions" in JSON format based on the prompt $P$ and the rendered image $I$.
\subsection{Stage 2: Direct Preference Optimization}
\label{sec:Stage 2}
The SFT stage instills the foundational "Generation" and "Critique" capabilities into our unified model, $\mathcal{M}_{\text{SFT}}$. In this stage, we exclusively target the "Generation" capability for DPO preference optimization. The core objective of this stage is to significantly enhance the model's "first-shot generation" quality, enabling it to produce more preferred results without iteration. We posit that a higher-quality initial draft is critical for the success of subsequent iterative refinement, as it provides a stronger starting point and may therefore reduce the total number of correction rounds required.

Training in this stage utilizes the generation preference data ($D_{\text{pref-G}}$) defined in the "Datasets" section. This dataset is constructed through a "generate-evaluate-pair" pipeline: for a given prompt $P_G$, we use $\mathcal{M}_{\text{SFT}}$ to generate $N$ candidate samples $\{S_i\}_{i=1}^N$. When automatically constructing preference pairs $(S_w, S_l)$, we adhere to the following rules: a renderable sample is always preferred over a non-renderable one; for two renderable samples, the sample with the higher expert score (with a score difference greater than $\delta$) is selected as the winner $S_w$.

We employ the DPO algorithm to optimize the SFT model $\mathcal{M}_{\text{SFT}}$ obtained from the first stage. DPO training requires both a policy model $\mathcal{M}_{\theta}$ (the model being optimized) and a reference model $\mathcal{M}_{\text{ref}}$ (whose parameters remain frozen). In our setup, $\mathcal{M}_{\text{SFT}}$ serves as the common starting point for both: we copy and freeze the weights of $\mathcal{M}_{\text{SFT}}$ to act as $\mathcal{M}_{\text{ref}}$, while simultaneously initializing $\mathcal{M}_{\theta}$ with the weights of $\mathcal{M}_{\text{SFT}}$. The training objective is to minimize the standard DPO loss $\mathcal{L}_{\text{DPO}}$ computed over $D_{\text{pref-G}}$:

\begin{equation*}
\resizebox{\columnwidth}{!}{
$\mathcal{L}_{\text{DPO}} = - \mathbb{E}_{(P_G, S_w, S_l)} \left[ \log \sigma \left( \beta \left( \log \frac{\mathcal{M}_\theta(S_w | P_G)}{\mathcal{M}_{\text{ref}}(S_w | P_G)} - \log \frac{\mathcal{M}_\theta(S_l | P_G)}{\mathcal{M}_{\text{ref}}(S_l | P_G)} \right) \right) \right]$
}
\end{equation*}

Upon completion of this stage, we define the optimized policy model $\mathcal{M}_{\theta}$ as our final unified model, $\mathcal{M}_{\text{Final}}$. Critically, because the SFT stage already functionally separates the "Generation" ($P_{\text{gen}}$) and "Critique" ($P, I$) capabilities within the model through different prompt formats, and the DPO stage is conducted only on "generation prompts" ($P_G$), this targeted preference tuning does not significantly disrupt the "Critique" capability that the SFT stage imparts. The resulting $\mathcal{M}_{\text{Final}}$ remains a single, unified model capable of both efficient generation and accurate critique.

\subsection{Introspective Refinement Loop}
\label{sec:Stage 3}
This represents the final application stage of our IntroSVG framework. In this stage, we use the final unified model $\mathcal{M}_{\text{Final}}$ obtained in the second stage. By switching prompt formats, the model seamlessly transitions between "Generator" and "Critic" roles to execute an iterative "Generate-Introspect-Refine" loop:

\begin{enumerate}
    \item \textbf{Generate}: The model $\mathcal{M}_{\text{Final}}$ receives a generation prompt. In the first round, it is the original user prompt $P_0$; in subsequent rounds, it is the "correction prompt" $P_{\text{gen}}$ constructed in the previous "Refine" step. The model performs the generation task, outputting SVG code $S_n$.
    
    \item \textbf{Critique}: $S_n$ is rendered into an image $I_n$. The \textit{same} $\mathcal{M}_{\text{Final}}$ model receives a "critique prompt" containing $P_0$ and the visual feedback $I_n$. It switches roles to perform introspection and outputs a structured evaluation $C_n$.
    
    \item \textbf{Termination Check}: If the score $\text{score}_n$ in $C_n$ meets a threshold or the maximum iteration count is reached, the loop terminates and outputs $S_n$.
    
    \item \textbf{Refine}: If the termination conditions are not met, the system constructs a new "correction prompt" $P_{\text{gen}} = \mathcal{T}(P_0, S_n, C_n)$.

    \item \textbf{Loop}: The system returns to the Generate step, feeding $P_{\text{gen}}$ back into the same model, $\mathcal{M}_{\text{Final}}$, to begin the next round of generation.
\end{enumerate}

The key to this architecture is its full utilization of the model's visual capabilities, enabling it to truly "see" the rendering feedback of its own work. This self-correction mechanism, grounded in authentic visual feedback, ultimately achieves an efficient, introspective, closed-loop self-correction using only a single model instance.

\begin{table}[!htbp]
\centering
\small
\caption{Impact Analysis of Data Standardization.}
\label{tab:evaluation_metrics}

\resizebox{\linewidth}{!}{
\begin{tabular}{l l l c c c c}
\toprule
\textbf{Dataset} & \textbf{Command} & \textbf{Precision} & \textbf{RSR\% $\uparrow$} & \textbf{FID $\downarrow$} &  \textbf{Aesthetic $\uparrow$} & \textbf{HPS $\uparrow$} \\
\midrule
$D_{base}$ & Mixed & Decimal & 68.41 & 121.50 & 4.2851 & 0.1815 \\
$D_{rel}$ & Relative & Integer & 97.16 & 37.34 & 4.7725 & 0.1846 \\
$D_{abs+decimal}$ & Absolute & Decimal & 95.89 & 34.62 & 4.7522 & 0.1877 \\
$D_{final}$ (Ours) & Absolute & Integer & \textbf{98.62} & \textbf{32.15} & \textbf{4.7963} & \textbf{0.1908} \\
\bottomrule
\end{tabular}%
} 
\end{table}

\begin{table*}[!htbp]
\centering
\caption{Comparison of different models across various metrics. The evaluation content includes image quality, semantic
information alignment effect, image style quality, and SVG code length, etc. Color convention: \textcolor{red}{best}, \textcolor{blue}{2nd-best}, and \textbf{3rd-best}.}
\label{tab:comparison-wide-resized}

\resizebox{0.95\textwidth}{!}{%
\begin{tabular}{lcccccc}
\toprule
\textbf{Method} & \textbf{Avg. Token $\downarrow$} & \textbf{RSR\% $\uparrow$} & \textbf{FID $\downarrow$} & \textbf{CLIP-T2I $\uparrow$} & \textbf{Aesthetic $\uparrow$} & \textbf{HPS $\uparrow$} \\
\midrule
GPT-4o \cite{achiam2023gpt} & 273.73 & \textcolor{red}{100} & 37.00 & 0.2748 & 4.4103 & 0.1941 \\
Gork-4 \cite{xai2025grok4} & 360.68 & \textcolor{red}{100} & 33.07 & 0.2717 & 4.4546 & 0.1944 \\
Claude 4.5 Sonnet \cite{anthropic2025claude4_5_system_card} & 439.42 & \textcolor{red}{100} & 39.67 & \textcolor{red}{0.2853} & 4.5724 & \textcolor{red}{0.1998} \\
Gemini 2.5 Pro \cite{gemini2025gemini2_5} & 356.00 & \textcolor{red}{100} & \textbf{30.52} &\textbf{ 0.2754} & 4.5854 & \textcolor{blue}{0.1981} \\
GPT-5 \cite{openai2025gpt5work} & 452.34 & \textcolor{red}{100} & 34.07 & \textcolor{blue}{0.2779} & 4.5232 & 0.1962 \\
\midrule
Qwen3-VL-30B-A3B-Instruct \cite{yang2025qwen3} & 410.77 & 96.35 & 38.68 & 0.2668 & 4.3911 & 0.1930 \\
InternVL3.5-38B-Instruct \cite{wang2025internvl3} & 529.21 & 91.07 & 57.49 & 0.2486 & 4.5157 & 0.1936 \\
Qwen2.5-VL-72B-Instruct \cite{bai2025qwen2} & 300.69 & 94.86 & 42.68 & 0.2533 & 4.4168 & 0.1906 \\
DeepSeek-R1 \cite{guo2025deepseek} & 314.45 & \textcolor{blue}{99.92} & 33.98 & 0.2734 & 4.5232 & 0.1962 \\
DeepSeek-V3.1 \cite{liu2024deepseek} & 367.58 & \textcolor{red}{100} & 36.18 & 0.2736 & 4.5539 & 0.1965 \\
\midrule
OmniSVG(Qwen2.5-3B-Instruct) \cite{yang2025omnisvg} & 2260.54 & 75.36 & 142.38 & 0.2297 & \textcolor{blue}{4.7232} & 0.1877 \\
SVGen(Qwen2.5-Coder-7B-Instruct) \cite{wang2025svgen} & 1531.42 & 84.64 & \textcolor{blue}{26.27} & 0.2339 & \textbf{4.5858} & 0.1916 \\
\midrule
IntroSVG(Ours) & 1707.77 & \textbf{99.26} & \textcolor{red}{26.18} & 0.2529 & \textcolor{red}{4.8894} & \textbf{0.1969} \\
\bottomrule
\end{tabular}
} 
\end{table*}

\section{Experiments}
\subsection{Effectiveness of Data Standardization}
To validate the necessity of our data standardization, we conduct an ablation study. Our core hypothesis is that a dataset with consistent syntax (absolute commands) and a concise representation (integer coordinates) significantly reduces the VLM's learning burden. To this end, we compare our final scheme, $\mathcal{D}_{final}$ (Absolute + Integer), against three key control groups: $\mathcal{D}_{base}$ (the raw baseline, Mixed + Decimal), $\mathcal{D}_{rel}$ (Relative + Integer), and $\mathcal{D}_{abs+decimal}$ (Absolute + Decimal). To ensure a fair comparison, we provide all SVGs as raw text sequences and train using Qwen2.5-VL-3B under identical configurations. The results are presented in Table~\ref{tab:evaluation_metrics}. These data clearly confirm our hypothesis: the $\mathcal{D}_{final}$ scheme, employing integer coordinates and absolute commands, significantly outperforms all control groups across all key metrics, strongly demonstrating the effectiveness of our standardization strategy.

\begin{figure*}[h] 
  \centering
  \includegraphics[width=\textwidth]{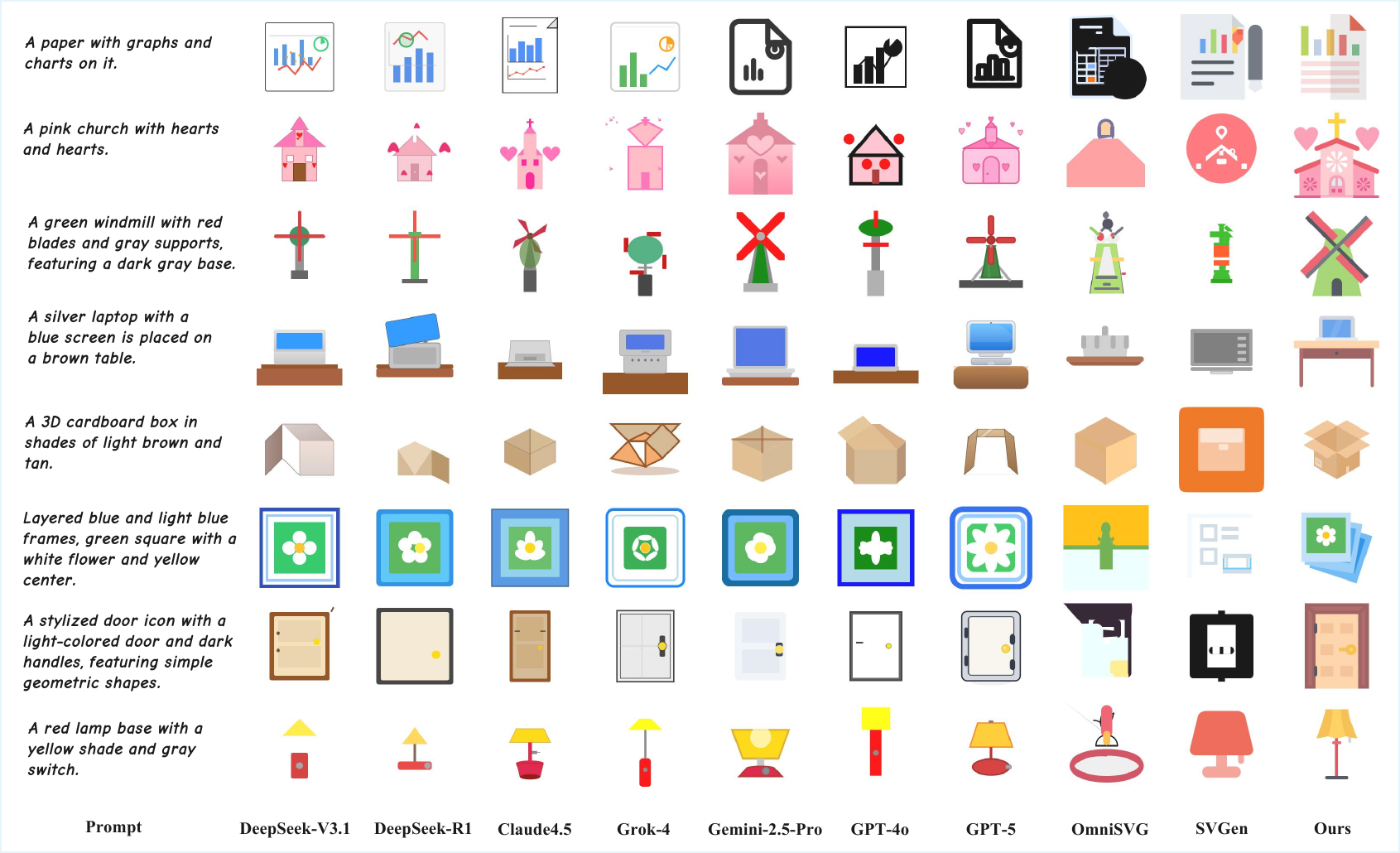} 
  \caption{Qualitative comparison between the proposed IntroSVG and other Text-to-SVG methods}
  \label{fig:resvis} %
\end{figure*}

\subsection{Evaluation Metrics}
We employ a suite of automatic evaluation metrics to comprehensively assess performance across code validity, visual quality, semantic alignment, and aesthetic appeal. Render Success Rate (RSR) measures the percentage of code that Cairosvg successfully renders. Avg.Token denotes the length of the generated SVG code after tokenization by the Qwen2.5 \cite{yang2024qwen2} tokenizer. Fréchet Inception Distance (FID) \cite{heusel2017gans} evaluates visual quality by comparing feature-space distributions; a lower score indicates that the generated images are closer to the real image distribution in visual quality and diversity. CLIPScore-T2I \cite{radford2021learning} computes the CLIP similarity between the image and the text to evaluate semantic alignment. For aesthetic appeal, we use both a pre-trained Aesthetic Score \cite{schuhmann2022} and a Human Preference Score (HPS) \cite{wu2023human} to better reflect overall human preferences.

\subsection{Baselines}
To comprehensively evaluate the performance of our proposed framework, we compare our model with the following categories of key models:

SOTA Domain-Specific Models: We select current mainstream models specifically trained for the SVG generation task, choosing their respective best-performing open-source versions for comparison: OmniSVG-3B \cite{yang2025omnisvg} and SVGen-Qwen2.5-Coder-7B-Instruct \cite{wang2025svgen}. We use their official implementations and evaluate them on a unified test set.

Closed-Source General-Purpose Models: We evaluate their ability to directly generate SVG code under zero-shot prompting. We test models including GPT-5 \cite{openai2025gpt5work}, Gemini 2.5 Pro \cite{gemini_20}, Grok-4 \cite{xai2025grok4}, among others.

Open-Source General-Purpose Models: We further divide this category into (1) Large Language Models, such as DeepSeek-R1 \cite{guo2025deepseek} and DeepSeek-V3.1 \cite{liu2024deepseek}; and (2) Large Vision-Language Models, such as Qwen2.5-VL-72B-Instruct \cite{yang2024qwen2}, Qwen3-VL-30B-A3B-Instruct \cite{yang2025qwen3}, and InternVL3.5-38B-Instruct \cite{wang2025internvl3}.

\subsection{Implementation Details}
All of our experiments are based on the Qwen2.5-VL-7B-Instruct \cite{yang2024qwen2}, a powerful VLM base model. All training is conducted on 8 NVIDIA A800 80GB GPUs. In the SFT stage (Sec. ~\ref{sec:Stage 1}), we use full-parameter fine-tuning, training on the mixed dataset $D_{\text{SFT}}$ for 3 epochs. We use the AdamW optimizer with a learning rate of $5 \times 10^{-5}$ and a cosine learning rate decay schedule. In the DPO stage (Sec. ~\ref{sec:Stage 2}), we perform DPO training for 3 epochs based on $\mathcal{M}_{\text{SFT}}$. The DPO learning rate is $5 \times 10^{-6}$, and the DPO $\beta$ parameter (KL divergence penalty coefficient) is set to 0.1. In the inference stage (Sec. ~\ref{sec:Stage 3}), for iterative inference, we set the maximum number of iterations to $N_{\text{max}}=3$ and the high-quality score threshold to $\tau=9.5$. The generator temperature is set to 0.5 for generation, while both the modification and critique processes use greedy decoding (temperature=0.0) to ensure deterministic results.

\section{Results and Analysis}
\subsection{Main Quantitative Analysis}
As Table~\ref{tab:comparison-wide-resized} shows, our IntroSVG demonstrates outstanding performance across all key metrics. Compared with existing SOTA domain-specific models (OmniSVG, SVGen), IntroSVG consistently outperforms them, achieving a near-perfect RSR (99.26\%)—far exceeding SVGen (84.64\%)—and ranking first in both visual quality (FID 26.18) and aesthetic score (Aesthetic 4.8894). More importantly, despite the powerful zero-shot capabilities of large general-purpose models, our 7B-parameter IntroSVG exhibits superior visual fidelity (FID 26.18 vs. 30.52) and aesthetic performance (Aesthetic 4.8894 vs. 4.5854) on this specialized task. This strongly demonstrates the effectiveness of our specialized training framework. The qualitative comparison in Figure~\ref{fig:resvis} also visually corroborates our quantitative advantages, showcasing IntroSVG's clear superiority in generating complex structures and faithfully adhering to text semantics compared with the baseline models.

\subsection{Ablation Studies}
We validate the necessity of each component in our framework in Table~\ref{tab:ablation_components}. The SFT stage serves as the foundation for performance improvement. With SFT training alone, the model's FID drops from 71.10 (Base Model) to 30.15, and the Aesthetic Score rises from 4.32 to 4.80. This demonstrates the critical role of our constructed mixed SFT dataset (containing $D_G^{\text{correction}}$ and $D_C$) in instilling generation and correction capabilities. Subsequently, the DPO stage further optimizes the model's "first-shot generation" quality. Without iteration (Iter 0), the FID decreases from 30.15 to 29.76, confirming that DPO successfully steers the model to prefer higher-quality initial drafts. Finally, activating the iterative loop is the key step to achieving SOTA performance, yielding a significant final boost in model performance, with the FID dropping to 26.18 and all metrics reaching their optimal values.

\begin{table*}[!htbp]
  \centering
  \small 
  \setlength{\tabcolsep}{5pt} 
  
  \caption{Ablation study on SFT data composition, DPO, and the iterative loop. This table shows the incremental contribution of each key component, starting from the raw base model.}
  \label{tab:ablation_components}
  
\begin{tabular}{l l c c c c c}
    \toprule
    \textbf{Model} & \textbf{Training Data} & \textbf{Iteration (Iter.)} & \textbf{FID ($\downarrow$)} & \textbf{CLIP-T2I ($\uparrow$)} & \textbf{Aesthetic ($\uparrow$)} & \textbf{HPS ($\uparrow$)} \\
    \midrule
    
    Qwen2.5-VL-7B (Base) & N/A (Zero-shot) & $\times$ & 71.10 & 0.2365 & 4.3240 & 0.1820 \\
    $\mathcal{M}_{\text{SFT}}$ & $D_{\text{SFT}}$ & $\times$ & 30.15 & 0.2472 & 4.8069 & 0.1910 \\
    $\mathcal{M}_{\text{Final}}$ & $D_{\text{SFT}} \cup D_{\text{pref-G}}$ & $\times$ & 29.76 & 0.2480 & 4.8372 & 0.1919 \\
    \textbf{$\mathcal{M}_{\text{Final}}$ (Iterative)} & \textbf{$D_{\text{SFT}} \cup D_{\text{pref-G}}$} & $\checkmark$ & \textbf{26.18} & \textbf{0.2529} & \textbf{4.8894 } & \textbf{0.1969} \\
    \bottomrule
  \end{tabular}
\end{table*}

\begin{figure}[h] 
  \centering
  \includegraphics[width=\linewidth]{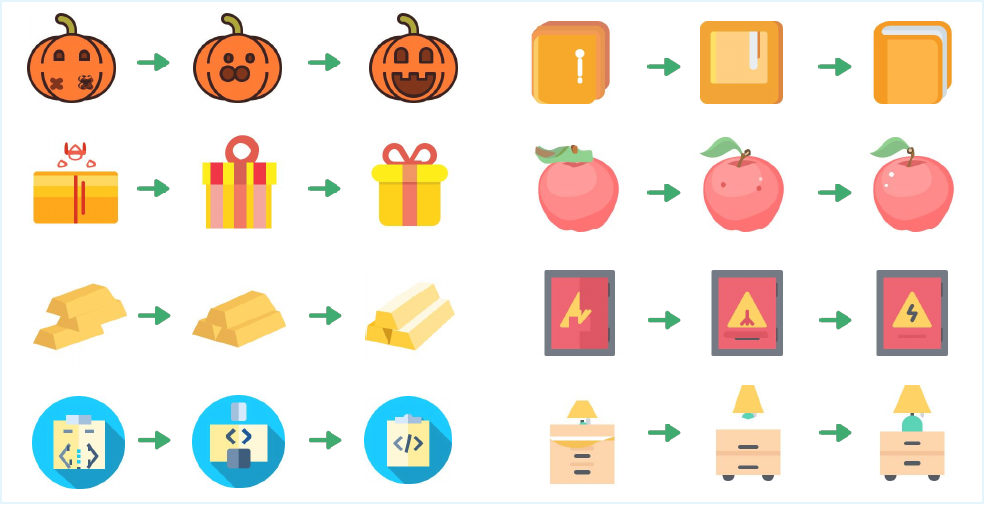} 
  \caption{Qualitative results of the iterative refinement loop.}
  \label{fig:example} 
\end{figure}

\subsection{Analysis of Iterative Refinement}
We further analyze the effectiveness of the introspective iterative loop. As shown in Table~\ref{tab:iteration_analysis}, this iterative process is both stable and efficient on our model: from Iter 0 to Iter 3, the FID continuously improves (decreasing from 29.76 to 26.18), while the Aesthetic and HPS scores steadily increase. This strongly demonstrates that our unified model is capable of fulfilling the dual roles of "critic" and "generator" and executing self-correction based on effective feedback. Moreover, we investigate the loop's generalization capability (Table~\ref{tab:generalizability_loop}). We apply this "generate-critique-refine" loop as a zero-shot prompting strategy to general-purpose models, such as GPT-4o and Grok-4. The results show that these models also achieve performance improvements (e.g., Grok-4's FID improves from 41.39 to 32.85). This indicates that the "introspective loop" is, in itself, a powerful inference framework with broad applicability.

\begin{table}[!htbp]
  \centering
  \footnotesize 
  \setlength{\tabcolsep}{5pt} 
  
  \caption{Quantitative analysis of the iterative refinement process, showing metric evolution from the initial draft up to $N_{\text{max}}=3$.}
  \label{tab:iteration_analysis}
  
  \begin{tabular}{l c c c c}
    \toprule
    \textbf{Iteration (N)} & \textbf{FID}($\downarrow$)   & \textbf{CLIP-T2I} ($\uparrow$)& \textbf{Aesthetic} ($\uparrow$) & \textbf{HPS} ($\uparrow$) \\
    \midrule
    
    0 (Initial Draft) & 29.76 & 0.2480 & 4.8372 & 0.1919 \\
    1 & 28.69 & 0.2511 & 4.8581 & 0.1931 \\
    2 & 27.65 & 0.2509 & 4.8774 & 0.1933 \\
    3 & \textbf{26.18} & \textbf{0.2529} & \textbf{4.8894} & \textbf{0.1969} \\
    \bottomrule
  \end{tabular}
\end{table}


\begin{table}[!htbp] 
  \centering
  \small
  \setlength{\tabcolsep}{3pt} 
  
  \caption{Evaluating the generalizability of the iterative loop. We test the "generate-critique-refine" cycle as a zero-shot prompting strategy on SOTA VLMs.}
  \label{tab:generalizability_loop}

  \resizebox{\linewidth}{!}{ 
    \begin{tabular}{l c c c c c}
      \toprule
      \textbf{Model} & \textbf{Iteration (N)} & \textbf{FID ($\downarrow$)} & \textbf{CLIP-T2I ($\uparrow$)} & \textbf{Aesthetic ($\uparrow$)} & \textbf{HPS ($\uparrow$)} \\
      \midrule
      
      GPT-4o & Iter 0& 37.80 & 0.2746 & 4.4121 & 0.1939 \\
      & Iter 1 & 36.92 & 0.2749 & 4.4477 & 0.1945 \\
      & Iter 2 & 36.62 & 0.2756 & 4.4633 & 0.1948 \\
      & Iter 3& \textbf{36.34} & \textbf{0.2771} & \textbf{4.4793} & \textbf{0.1949} \\
      \midrule
      
      Grok-4 & Iter 0 & 41.39 & 0.2696 & 4.4551 & 0.1941 \\
      & Iter 1 & 39.58 & 0.2714 & 4.4758 & 0.1942 \\
      & Iter 2 & 36.40 & 0.2715 & 4.4854 & 0.1940 \\
      & Iter 3& \textbf{32.85} & \textbf{0.2725} & \textbf{4.4855} & \textbf{0.1941} \\
      \midrule
      
      GPT-5 & Iter 0& 35.87 & 0.2756 & 4.6069 & 0.1984 \\
      & Iter 1 & 34.91 & 0.2858 & 4.6156 & 0.1985 \\
      & Iter 2 & 33.95 & \textbf{0.2858} & \textbf{4.6324} & 0.1987 \\
      & Iter 3 & \textbf{32.68} & 0.2779 & 4.6264 & \textbf{0.1999} \\
      
      \bottomrule
    \end{tabular}
  }
\end{table}

\section{Conclusions}
This study addresses a critical limitation of existing Text-to-SVG methods: a limited awareness of the visual characteristics of rendered outputs and an insufficient ability to self-correct. We introduce IntroSVG, an introspective framework for SVG generation. Our core contribution is to implement a unified Vision-Language Model that concurrently serves as both the "Generator" and the "Critic" within a closed-loop framework. Through multi-task SFT that leverages failure samples for error correction, together with DPO preference alignment, the model learns to assess its own visual outputs.

During inference, the model employs a "generate-critique-refine" iterative loop to autonomously refine imperfect drafts. Experimental results show that IntroSVG achieves state-of-the-art (SOTA) performance on key metrics, including visual quality (FID) and aesthetic scores, and substantially outperforms both existing domain-specific and large general-purpose models. Ablation studies further validate the effectiveness of each component of the framework—SFT, DPO, and the iterative loop—and underscore the importance of integrating explicit visual feedback into the generative process. Looking ahead, we plan to extend this autonomous loop into an interactive editing tool, where human instructions can act as an "external critique" signal to enable more controllable, human-in-the-loop optimization.

\section*{Acknowledgements}

This work was supported in part by grants from the National Natural Science Foundation of China (62306241 \& U62576284).


\bibliographystyle{ieeenat_fullname}
\bibliography{reference}

\clearpage
\setcounter{page}{1}
\maketitlesupplementary

\section{Data processing pipeline}
\label{sec:rationale}
\subsection{SVG-related datasets and benchmarks}
The development of data resources in the field of vector graphics generation exhibits a dual-track evolutionary trajectory: regarding training data, there has been a shift from large-scale unsupervised collection toward a refined approach focused on attribute enhancement and reasoning guidance; concurrently, the evaluation system has matured, establishing comprehensive benchmarks that encompass multi-dimensional, multi-format, and fine-grained editing capabilities.

\textbf{Large-scale Foundation Datasets:} To address the scarcity of training data, early research focused on constructing million-scale datasets to cover a wide distribution of graphics. StarVector introduced SVG-Stack, containing 2.1 million real code samples from GitHub . It retains diagrams and complex primitives (such as circles and polygons), making it one of the datasets with the most authentic code structures . OmniSVG constructed MMSVG-2M (2 million samples), innovating by introducing high-complexity anime characters (20\%) and illustrations, and supplementing complex data through diffusion model generation and vectorization techniques . IconShop, based on the FIGR-8-SVG dataset (1.5 million monochrome icons), utilized ChatGPT to expand discrete keywords into natural language descriptions, establishing an early large-scale benchmark for text-to-icon generation . UniSVG (525k) further broke down task barriers by integrating image generation, text generation, and graphic understanding into a unified, cleaned dataset, supporting the all-around fine-tuning of Multimodal Large Language Models (MLLMs).

\textbf{Attribute-Enhanced and Colored Datasets:} Addressing the limitations of early data being mostly monochrome or simple outlines, subsequent datasets emphasized enhancing visual richness and structural attributes. SVGBuilder proposed ColorSVG-100K, the first large-scale dataset specifically for colored SVGs (100,000 items), filling the gap in color information in previous datasets . SVGen constructed SVG-1M (1 million), innovatively grading data complexity based on color and command count (Easy/Difficult) to support curriculum learning for models . LLM4SVG, through 250k SVGs and 580k instruction pairs, emphasized treating SVG as semantic tokens for structured understanding.

\textbf{Reasoning and Process-Oriented Datasets:} With the improvement of model capabilities, data construction began to focus on the logic and process behind generation. Reason-SVG's SVGX-DwT-10k contains 10,000 curated samples, each equipped with detailed "Chain-of-Thought (CoT)" annotations, recording the complete design flow from conceptual design to coordinate calculation . SVGThinker built a serialized dataset containing 270,000 samples, generating intermediate state images and descriptions corresponding to each step's instruction by reconstructing the SVG tree structure, training the model to understand the logical order of drawing.

\textbf{Multi-format Benchmarks:} Evaluation benchmarks have gradually expanded from single generation tasks to understanding and fine-grained editing. VGBench is a broad benchmark that evaluates not only SVG but also TikZ and Graphviz formats, containing 4,279 understanding Q/A pairs and 5,845 generation samples, aiming to assess the general capability of LLMs across different vector languages. SVGEditBench V2 focuses on instruction-level editing, containing 1,683 "Original Image - Instruction - Target Image" triplets built from Emoji datasets, specifically testing the model's ability to modify images (e.g., changing color, rotating) while maintaining the original structure. SVGenius further introduced complexity stratification (Easy/Medium/Hard), comprehensively covering tasks such as understanding, bug fixing, code optimization, and style transfer, providing a more discriminative capability assessment.

To demonstrate the advantages of our data processing, we provide a detailed comparison between our IntroSVG dataset and the original source datasets (OmniSVG, LLM4SVG, SVGen) in Table~\ref{tab:svg_projects}.

\begin{table*}[t]
\centering
\small
\caption{Comparison of statistics between our IntroSVG dataset and the source datasets.} \label{tab:svg_projects}

\resizebox{\textwidth}{!}{%
\begin{tabular}{l c c c c c c}
\toprule
\textbf{Project} & \textbf{SVG Num} & \textbf{Num of multicolor SVG} & \textbf{Command Type} & \textbf{Precision} & \textbf{Unified Viewbox} & \textbf{Source} \\
\midrule
OmniSVG & 500k & 110k (22\%) & Absolute & Decimal & 200, 200 & Iconfont, Iconscout \\
LLM4SVG & 250k & 80k (32\%) & Relative & Decimal & 128, 128 & TwEmoji, NotoEmoji, Reshot, SVGRepo \\
SVGGen & 500k & 140k (28\%) & Absolute & Integer & 1024, 1024 & Iconfont \\
IntroSVG & 200k & 200k (100\%) & Absolute & Integer & 200, 200 & All of the above \\
\bottomrule
\end{tabular}%
}
\end{table*}

\begin{figure*}[htbp]
    \centering
    \includegraphics[width=\textwidth]{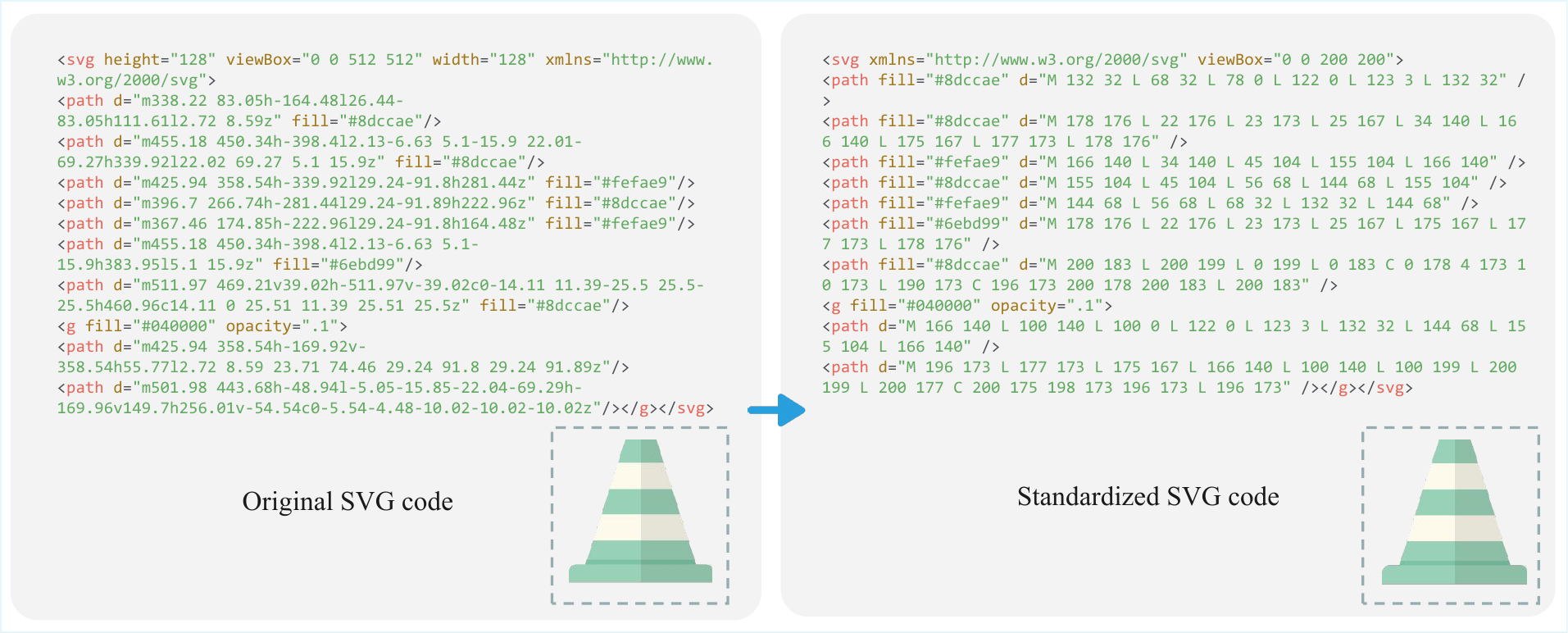}
    \caption{Visual and code comparison before and after data standardization.}
    \label{fig:processing_comparison}
\end{figure*}

\subsection{Data Collection and Preprocessing}
\textbf{Data Sources and Motivation:} 
Although existing large-scale SVG datasets are abundant, utilizing them directly for training presents significant challenges. First, large-scale web-crawled datasets often exhibit severe sample redundancy and high similarity. This not only wastes computational resources during training but also exacerbates the risk of model overfitting. Second, data from diverse sources displays significant heterogeneity in formatting specifications, characterized by inconsistent \texttt{viewBox} dimensions, varying coordinate precision (including the number of decimal places), and the mixed usage of relative and absolute path commands. This distributional inconsistency substantially increases the difficulty for models to learn underlying geometric patterns.

To construct a high-quality, standardized colored vector icon dataset that supports complex SVG generation while mitigating overfitting risks, we integrated three mainstream open-source datasets: LLM4SVG, OmniSVG, and SVGen. These datasets encompass rich icon semantics and diverse visual styles, providing a solid foundation for our training.

\textbf{Data Cleaning and Standardization Pipeline:} 
To ensure data uniformity and high quality, we designed and implemented a rigorous filtering and standardization pipeline, detailed as follows:

\begin{table}[!htbp]
    \centering
    \renewcommand{\arraystretch}{1.5} 
    \setlength{\tabcolsep}{4pt} 
    \begin{tabularx}{\columnwidth}{>{\centering\arraybackslash}m{1.8cm} >{\centering\arraybackslash}m{1.2cm} >{\centering\arraybackslash}m{2.2cm} >{\centering\arraybackslash}m{2cm}}
    \hline
    Name & Symbol & Arguments & Visualization \\ 
    \hline
    Move To & M & $(x_1, y_1), (x_2, y_2)$ & \includegraphics[height=1cm,width=1.5cm,keepaspectratio]{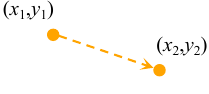} \\
    \hline
    Line To & L & $(x_1, y_1), (x_2, y_2)$ & \includegraphics[height=1cm,width=1.5cm,keepaspectratio]{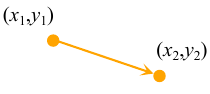} \\
    \hline
    Cubic Bézier & C & \begin{tabular}{@{}c@{}}$(x_1, y_1), (q_1^x, q_1^y),$ \\ $(q_2^x, q_2^y), (x_2, y_2)$\end{tabular} & \includegraphics[height=1cm,width=1.5cm,keepaspectratio]{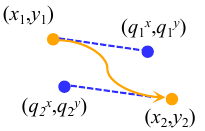} \\
    \hline
    Elliptical Arc & A & \begin{tabular}{@{}c@{}}$(r_x, r_y),\varphi,f_L ,$ \\ $ f_S, (x, y)$\end{tabular} & \includegraphics[height=1.75cm,width=1.75cm,keepaspectratio]{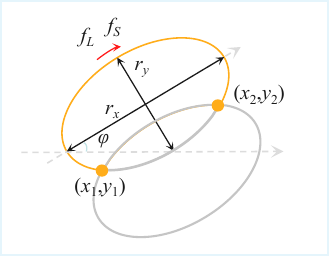} \\
    \hline
    ClosePath & Z & Close the path & - \\
    \hline
    \end{tabularx}
    \caption{List of the simplified SVG command vocabulary.}
    \label{tab:svg_commands}
\end{table}

\noindent\textbf{Quality Filtering:}
    \begin{itemize}
        \item \textit{Removal of Monochrome Samples:} To focus on generating colored icons with rich visual information, we excluded pure monochrome samples.
        \item \textit{Removal of Invalid Samples:} We filtered out corrupted files that could not be rendered by standard rendering engines (CairoSVG).
        \item \textit{Sequence Length Constraint:} We removed samples with token sequence lengths exceeding 8000 to ensure memory efficiency during training and stability during inference.
    \end{itemize}

\noindent\textbf{Geometric Normalization:}
    \begin{itemize}
        \item \textit{Unified Canvas:} The \texttt{viewBox} attributes of all SVG samples were rescaled and normalized to ``0 0 200 200'', eliminating scale ambiguities caused by differing canvas dimensions.
        \item \textit{Primitive Unification:} All basic shape elements (e.g., \texttt{<rect>}, \texttt{<circle>}, \texttt{<ellipse>}) were converted into generic \texttt{<path>} elements.
        \item \textit{Command Standardization:} We converted all relative commands in path data to absolute commands and retained only five core instruction types: Move (\textbf{M}), Line (\textbf{L}), Cubic B\'{e}zier Curve (\textbf{C}), Elliptical Arc (\textbf{A}), and Close Path (\textbf{Z}), thereby simplifying the vocabulary and unifying the semantic space, as defined in Table~\ref{tab:svg_commands}.
    \end{itemize}

\noindent\textbf{Numerical and Format Standardization:}
    \begin{itemize}
        \item \textit{Integer Coordinates:} All floating-point coordinates in paths were rounded to integers. This step significantly reduced token length while maintaining visual fidelity, lowering the difficulty for the model to predict continuous values.
        \item \textit{Attribute Reordering:} We standardized SVG file headers and enforced a specific attribute order within each \texttt{<path>} tag, placing the \texttt{fill} (color) attribute before the \texttt{d} (path data) attribute. This design aims to guide the model to plan the color style first before generating specific geometric paths, establishing a consistent generation sequence pattern.
    \end{itemize}

\begin{figure*}[b] 
    \centering
    \begin{tcolorbox}[
        colframe=blue!30!black,
        colback=blue!5!white,
        boxrule=1pt,
        arc=4pt,
        outer arc=4pt,
        boxsep=5pt,
        left=6pt,
        right=6pt,
        top=6pt,
        bottom=6pt,
        fonttitle=\bfseries,
        title=Prompt: GPT-4o image description,
        ]
    \small
    \textbf{Role Play}
    
    You are a world-class SVG design master and code engineer, possessing exceptional design aesthetics and a profound understanding of SVG code structure. Your task is to act as a mentor reviewing an SVG work. Your core expertise lies in the ability to reverse-engineer the underlying SVG paths, shapes, and layer structure from a rendered image and provide code-level optimization advice.

    \vspace{0.5em} 
    \textbf{Task Context}
    
    A draft has been generated based on an original design requirement, and a high-quality reference version designed by a human expert is also provided. Both images are rendered from SVG code. Your task is to compare these two works and generate a professional review report for the draft. The report should guide on how to modify the SVG code to learn from and improve towards the reference version, while also considering the draft's consistency with the original design prompt.

    \vspace{0.5em}
    \textbf{Contextual Information}
    \begin{enumerate}
        \item \textbf{Original Prompt:} \texttt{\{original\_prompt\}} 
        \item The draft image is the first input image, used to present the generated design draft.
        \item The high-quality reference image is the second input image, used to present the visual effect of the reference design.
    \end{enumerate}

    \textbf{Your Core Task \& Scoring Guide}
    
    Please carefully observe and compare the first image (the draft) with the second image (the high-quality reference version), while also considering the requirements of the original design prompt. Output a detailed review report in JSON format. Your sole task is to generate this report; do not generate any SVG code.

    \vspace{0.5em}
    \textbf{Important Scoring Instructions}
    \begin{itemize}
        \item When the draft is highly similar to the reference version and meets the requirements of the original design prompt, please give a high score (e.g., 9.0-10.0). In the suggestions, do not provide modification guidance, only an affirmative output.
        \item When there are significant differences between the draft and the reference version, or deviations from the original design prompt, please give a reasonable mid-to-low score based on the severity of the differences and the visual quality. Provide 2 to 4 specific, actionable SVG modification suggestions.
    \end{itemize}

    \textbf{JSON Output Format}
    
    The JSON object must contain the following three fields:
    \begin{itemize}
        \item \texttt{"score"}: A float, ranging from 0.0 to 10.0, for the overall rating of the draft.
        \item \texttt{"critique"}: A comprehensive evaluation explaining the draft's adherence to the original design prompt and pointing out the main differences and shortcomings in aesthetics, color, and geometric construction compared to the reference version.
        \item \texttt{"suggestions"}: If the draft is very close to the reference version and aligns with the original prompt, use an affirmative statement such as "The overall quality is excellent, no changes needed." Otherwise, provide specific SVG modification suggestions.
    \end{itemize}
    
    \textit{Goal: Create robust SVG representations for research purposes.}
    \end{tcolorbox}
    \caption{Prompt template for constructing the Critique Dataset.}
    \label{fig:desc_gen_pro_4o}
\end{figure*}

The impact of this standardization pipeline is visually demonstrated in Figure~\ref{fig:processing_comparison}. The process effectively transforms raw, mixed-format code into a clean, unified representation on a $200\times200$ canvas, significantly reducing sequence complexity. Following this rigorous processing pipeline, we ultimately filtered and processed approximately \textbf{200,000} high-quality (Text prompt, SVG code) pairs, forming the basis for the experiments in this paper.

\section{Data Construction}
In this section, we provide a granular description of the data synthesis pipeline used to construct the training sets for both the Supervised Fine-Tuning (SFT) and Direct Preference Optimization (DPO) stages. The construction process leverages a "Generator-Critic" loop involving GPT-4o to synthesize high-quality instruction-following data.

\subsection{SFT Dataset Construction}
The SFT dataset $D_{\text{SFT}}$ is a mixture of three distinct subsets: $D_{\text{SFT}} = D_G^{\text{direct}} \cup D_G^{\text{correction}} \cup D_C$. Each subset targets a specific capability of the unified model.

\subsubsection{Foundational Generation Data ($D_G^{\text{direct}}$)}
This dataset instills the core capability of translating text to SVG code.

\noindent\textbf{Source:} The 200k standardized samples from our data cleaning pipeline.

\noindent\textbf{Structure:} Direct pairs of $(X, Y)$, where $X$ is the descriptive textual prompt and $Y$ is the canonical SVG code.

\subsubsection{Correction and Critique Data Synthesis}
To equip the model with self-correction and self-critique capabilities, we constructed synthetic datasets $D_G^{\text{correction}}$ and $D_C$. The synthesis pipeline is as follows:

\begin{itemize}
    \item \textbf{Draft Generation:} We first trained a temporary model (warm-up) on $D_G^{\text{direct}}$ for one epoch. We then selected 50,000 prompts from the validation set and generated initial SVG drafts using this model. These drafts intentionally contain imperfections (e.g., geometric distortions, color mismatches) typical of early-stage training.
    
    \item \textbf{Expert Annotation (GPT-4o):} We employed GPT-4o as an "Teacher VLM" to evaluate these drafts. We utilized the structured prompt shown in Figure~\ref{fig:desc_gen_pro_4o}. For each triplet of (Original Prompt, Draft Image, Reference Image), GPT-4o generated a JSON response containing:

        \textbf{score}: A quantitative quality assessment (0.0-10.0).
        
         \textbf{critique}: A textual analysis of flaws in geometry and aesthetics.
         
         \textbf{suggestions}: Actionable advice for code modification.

    \item \textbf{Dataset Formulation:} Based on the expert feedback, we constructed the specific training samples:
    
         \textbf{Critique Dataset ($D_C$):} Inputs are the prompt and the rendered draft image; the target output is the expert's JSON critique.
         
         \textbf{Correction Dataset ($D_G^{\text{correction}}$):} Inputs are the complex prompt (containing the original prompt, the flawed draft code, and the expert's critique); the target output is the high-quality ground truth SVG from $D_G^{\text{direct}}$.
\end{itemize}

\begin{figure}[htbp]
    \centering
    \begin{tcolorbox}[
        colframe=blue!30!black,
        colback=blue!5!white,
        boxrule=1pt,
        arc=4pt,
        outer arc=4pt,
        boxsep=5pt,
        left=6pt,
        right=6pt,
        top=6pt,
        bottom=6pt,
        fonttitle=\bfseries,
        title=Prompt: SVG Quality Scoring, 
        ]
    \small
    \textbf{Task}
    
    I used this prompt: \texttt{\{original\_prompt\}}
    
    Rate the 5 SVG images I'm uploading from 1-100 based on that prompt. Ensure the scores are differentiated.

    \vspace{0.5em}
    \textbf{Evaluate based on:}
    \begin{itemize}
        \item \textbf{Prompt Adherence:} How well the image matches the prompt's elements and mood.
        \item \textbf{Visual Aesthetics:} Color, composition, and visual impact.
        \item \textbf{Execution Quality:} Creativity and technical quality (clean SVG, no flaws).
    \end{itemize}

    \vspace{0.5em}
    \textbf{Output Format}
    
    Provide only JSON in this exact format. No other text.

    \vspace{0.5em}
    \texttt{\{ \\
    \hspace*{1em} "image\_1\_score": [Score], \\
    \hspace*{1em} "image\_2\_score": [Score], \\
    \hspace*{1em} "image\_3\_score": [Score], \\
    \hspace*{1em} "image\_4\_score": [Score], \\
    \hspace*{1em} "image\_5\_score": [Score] \\
    \}}
    \end{tcolorbox}
    \vspace{-10pt}
    \caption{Prompt used for scoring generated SVG candidates}
    \label{fig:svg_scoring_prompt}
\end{figure}

\subsection{DPO Preference Dataset Construction}
For the DPO stage, we constructed a preference dataset $D_{\text{pref-G}}$ to optimize the model's first-pass generation quality. This process involves sampling, scoring, and pair selection.

\subsubsection{Candidate Sampling}
We selected a diverse set of 10,000 prompts. Using the converged SFT model ($M_{\text{SFT}}$), We generated $N=5$ distinct candidate SVGs for each prompt with a temperature of 0.9, resulting in a pool of 50,000 candidate samples.

\subsubsection{Automated Scoring}
We employed GPT-4o as an automated evaluator to score each candidate. The prompt used for this process is illustrated in Figure~\ref{fig:svg_scoring_prompt}. The scoring criteria explicitly cover:
\begin{itemize}
    \item \textbf{Prompt Adherence:} Alignment with the user's text instructions.
    \item \textbf{Visual Aesthetics:} Color harmony and geometric balance.
    \item \textbf{Execution Quality:} Syntactic correctness and renderability.
\end{itemize}

\subsubsection{Preference Pair Construction}
To construct the training triplets $(Prompt, S_w, S_l)$, where $S_w$ is the winning sample and $S_l$ is the losing sample, we applied the following hierarchical rules:

\begin{itemize}
    \item \textbf{Rule 1: Render-Success Priority.} A renderable SVG is strictly preferred over a non-renderable one (e.g., one with syntax errors or invalid paths). If Candidate A renders and Candidate B fails, then $S_w=A, S_l=B$.
    \item \textbf{Rule 2: High-Score Priority.} For two renderable candidates, we compare their GPT-4o scores. To ensure distinct separability and avoid noise from similar-quality samples, we enforced a margin $\delta$. If $Score(A) - Score(B) > \delta$, then $S_w=A, S_l=B$.
\end{itemize}

This results in the dataset $D_{\text{pref-G}}$, ensuring DPO improves both the model's syntactic robustness and visual aesthetics.

\begin{figure*}[t]
    \centering
    
    \newcommand{\sectiontitle}[1]{\textbf{\textcolor{blue!40!black}{\sffamily #1}}}
    
    \newtcolorbox{examplebox}{
        enhanced,
        colback=white,          
        colframe=gray!20,       
        boxrule=0.5pt,          
        left=2pt, right=2pt, top=4pt, bottom=4pt,
        sidebyside,             
        sidebyside align=center,
        righthand width=0.14\textwidth, 
        sidebyside gap=10pt,    
        code={\small}           
    }

    \begin{tcolorbox}[
        enhanced,
        title={\bfseries \large Prompt and Critique Examples}, 
        coltitle=black,
        colbacktitle=gray!10,   
        colframe=black,
        colback=white,
        boxrule=1pt,            
        arc=2pt,                
        drop fuzzy shadow,      
        left=4pt, right=4pt, top=4pt, bottom=4pt
    ]

    \begin{examplebox}
        \sectiontitle{Original prompt:}\\
        An umbrella with a yellow and blue color.
        
        \vspace{3pt}
        \sectiontitle{Critique \& Refine:}
        \begin{itemize}[leftmargin=*, nosep] 
            \item \textbf{Score:} 4.5
            \item \textbf{Critique:} The draft image significantly deviates from the reference version and the original design prompt. The reference image showcases a smooth, aesthetically pleasing umbrella with a yellow and blue color scheme, whereas the draft image presents a pixelated, simplistic version that lacks the smooth curves.
            \item \textbf{Suggestions:} 
            \begin{enumerate}[nosep, leftmargin=1em, label=\arabic*.]
                \item Smooth out the umbrella's canopy by using curved paths instead of blocky, pixelated shapes to match the reference's aesthetic.
                \item Adjust the color distribution to include both yellow and blue in a balanced manner, ensuring the yellow is prominent.
                \item Refine the handle to have a smooth, curved appearance rather than a straight, pixelated form.
                \item Ensure the overall proportions and symmetry of the umbrella are consistent with the reference image.
            \end{enumerate}
        \end{itemize}
        \tcblower 
        \centering
        \includegraphics[width=\linewidth]{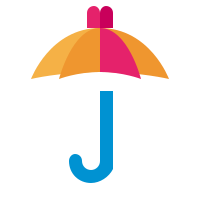}
    \end{examplebox}
    
    \vspace{1pt} 

    \begin{examplebox}
        \sectiontitle{Original prompt:}\\
        A white and blue cloud sits beneath a two-tone blue square with vibrant yellow and orange forming the background.
        
        \vspace{4pt}
        \sectiontitle{Critique \& Refine:}
        \begin{itemize}[leftmargin=*, nosep]
            \item \textbf{Score:} 6.5
            \item \textbf{Critique:} The draft image captures the essence of the original design prompt by featuring a cloud beneath a two-tone blue square. However, there are notable differences compared to the reference version. The draft uses a more simplified cloud design and lacks the subtle color variations present in the reference.
            \item \textbf{Suggestions:} 
            \begin{enumerate}[nosep, leftmargin=1em, label=\arabic*.]
                \item Add the small colored squares at the top of the blue square to match the reference design.
                \item Adjust the color tones of the two-tone blue square to be more muted, aligning with the reference.
                \item Introduce subtle color variations in the cloud to enhance its visual depth and match the reference.
                \item Improve the integration of the cloud and square into the background for a more cohesive composition.
            \end{enumerate}
        \end{itemize}
        \tcblower
        \centering
        \includegraphics[width=\linewidth]{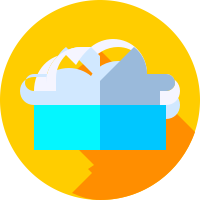}
    \end{examplebox}

    \vspace{1pt}

    \begin{examplebox}
        \sectiontitle{Original prompt:}\\
        A cheerful yellow sun with a smiling face and closed eyes radiates joy and warmth.
        
        \vspace{3pt}
        \sectiontitle{Critique \& Refine:}
        \begin{itemize}[leftmargin=*, nosep]
            \item \textbf{Score:} 6.5
            \item \textbf{Critique:} The draft image captures the essence of a cheerful sun with a smiling face and closed eyes, aligning with the original design prompt. However, there are notable differences compared to the reference version. The draft uses a more geometric and angular style for the sun's rays, whereas the reference has smoother, more rounded rays. The facial features in the draft are simplistic and lack the expressive detail seen in the reference.
            \item \textbf{Suggestions:} 
            \begin{enumerate}[nosep, leftmargin=1em, label=\arabic*.]
                \item Modify the sun's rays to have a smoother, more rounded appearance to match the softer aesthetic of the reference image.
                \item Enhance the facial features by adding more detail and expression, particularly in the eyes and mouth, to better convey joy and warmth.
                \item Adjust the color palette to increase contrast and vibrancy, ensuring the sun's face stands out more distinctly from its rays.
                \item Consider refining the overall symmetry and balance of the design to align more closely with the reference's harmonious composition.
            \end{enumerate}
        \end{itemize}
        \tcblower
        \centering
        \includegraphics[width=\linewidth]{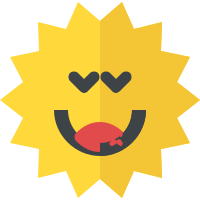} 
    \end{examplebox}

    \end{tcolorbox}
    \vspace{-10pt}
    \caption{Examples of critique data generated by GPT-4o}
    \label{fig:critique_examples}
\end{figure*}

\begin{figure*}[!t]
    \centering
    
    \begin{tcolorbox}[
        enhanced,
        colframe=black,
        colback=white,
        coltitle=black,
        boxrule=1.2pt,
        arc=0pt, outer arc=0pt,
        title=Training Data Samples,
        fonttitle=\bfseries,
        colbacktitle=titlebg,
        titlerule=0pt,
        top=6pt, bottom=6pt, left=6pt, right=6pt
    ]
    \small 

        \begin{innerbox}
            \role{User} Please generate an SVG icon that meets the following description: A brown wallet with a light beige accent and a yellow button represents saving or spending money.
            
            \vspace{0.4em}
            \role{Assistant}
            \begin{codefont}
            <svg xmlns="http://www.w3.org/2000/svg" viewBox="0 0 200 200">\\
            \hspace*{1em}<path fill="\#9A8984" d="M24 55 L5 41 L5 170 C5 184 16 195 30 195 L195 195..." />\\
            \hspace*{1em}<path fill="\#BCB0AD" d="M165 5 L30 5 C16 5 5 16 5 30 C5 37 9 46 24 55..." />\\
            \hspace*{1em}<path fill="\#FEDEA1" d="M175 125 A15 15 0 1 1 145 125 A15 15 0 1 1 175..." />\\
            \hspace*{1em}<path fill="\#573A32" d="M160 145 C171 145 180 136 180 125 C180 114 171 105..." />\\
            \hspace*{1em}<path fill="\#573A32" d="M190 50 L170 50 L170 10 C170 4 166 0 160 0 L30 0..." />\\
            </svg>
            \end{codefont}
        \end{innerbox}

        \vspace{0.6em}

        \begin{innerbox}
            \role{User} Please analyze all the information provided below and generate a final, high-quality SVG code. The original design goal was: "A brown wallet with a light beige accent and a yellow button represents saving or spending money."; a draft SVG code that needs improvement is as follows:
            
            \vspace{0.2em}
            \begin{codefont}
            <svg xmlns="http://www.w3.org/2000/svg" viewBox="0 0 200 200">\\
            \hspace*{1em}<path fill="\#E8CD97" d="M166 156 L70 156 C66 156 62 152 62 148 L62 85..." />\\
            \hspace*{1em}<path fill="\#472F17" d="M166 161 L33 161 C25 161 18 154 18 146 L18 53..." />\\
            \hspace*{1em}<path fill="\#472F17" d="M184 121 L126 121 C112 121 101 110 101 96 L101 95..." />\\
            \hspace*{1em}<path fill="\#472F17" d="M117 96 A8 8 0 1 0 133 96 A8 8 0 1 0 117 96 Z" />\\
            \hspace*{1em}<path fill="\#472F17" d="M37 95 L18 95 L18 74 L37 74 L37 95 Z M18 85..." />\\
            </svg>
            \end{codefont}
            
            \vspace{0.2em}
            An expert critique of this draft is: "The draft image presents a wallet design that partially adheres to the original prompt... The draft's geometric construction is more abstract and less realistic... The draft lacks the depth and shading..."
            
            Specific modification suggestions: 1. Adjust the wallet's shape... 2. Enhance the shading... 3. Increase the prominence of the button... 4. Consider using a more muted brown tone...
            
            \vspace{0.4em}
            \role{Assistant}
            \begin{codefont}
            <svg xmlns="http://www.w3.org/2000/svg" viewBox="0 0 200 200">\\
            \hspace*{1em}<path fill="\#9A8984" d="M24 55 L5 41 L5 170 C5 184 16 195 30 195 L195 195..." />\\
            \hspace*{1em}<path fill="\#BCB0AD" d="M165 5 L30 5 C16 5 5 16 5 30 C5 37 9 46 24 55..." />\\
            \hspace*{1em}<path fill="\#FEDEA1" d="M175 125 A15 15 0 1 1 145 125 A15 15 0 1 1 175..." />\\
            \hspace*{1em}<path fill="\#573A32" d="M160 145 C171 145 180 136 180 125 C180 114 171 105..." />\\
            \hspace*{1em}<path fill="\#573A32" d="M190 50 L170 50 L170 10 C170 4 166 0 160 0 L30 0..." />\\
            </svg>
            \end{codefont}
        \end{innerbox}

        \vspace{0.6em}

        \begin{innerbox}
            \role{User} You are a professional SVG design critic. Please analyze the input AI-generated SVG draft \textless \textbf{image}\textgreater according to the "Original Design Prompt". Original Design Prompt: "A brown wallet with a light beige accent and a yellow button..." Your task is to output a structured critique report in JSON format (score, critique, suggestions).

            \vspace{0.4em}
            \role{Assistant}
            \begin{codefont}
            \{\\
            \hspace*{1em}"score": 5.5,\\
            \hspace*{1em}"critique": "The draft image presents a wallet design that partially adheres to the original prompt. The color scheme in the draft uses a brown and beige palette, which aligns with the prompt, but the execution differs significantly from the reference. The draft's geometric construction is more abstract and less realistic compared to the reference...",\\
            \hspace*{1em}"suggestions": "1. Adjust the wallet's shape to have rounded corners and a more traditional wallet appearance, similar to the reference. 2. Enhance the shading and depth to give the wallet a more three-dimensional look. 3. Increase the prominence of the button by making it larger and more central... 4. Consider using a more muted brown tone..."\\
            \}
            \end{codefont}
        \end{innerbox}

    \end{tcolorbox}
    \vspace{-10pt}
    \caption{Training Data Formats for Different Capabilities. Top: Generation data $D_G^{direct}$. Middle: Correction data $D_G^{correction}$ with draft and critique inputs. Bottom: Critique data $D_C$} 
    \label{fig:dataset_samples}
\end{figure*}

\section{Additional Evaluation}
\label{sec:additional_eval}

In this section, we provide additional experimental analyses to further validate the effectiveness and reliability of the proposed IntroSVG framework. Specifically, we report results on an additional benchmark dataset, analyze the computational efficiency of the iterative generation strategy, and conduct human evaluation to assess perceptual quality.

\subsection{Evaluation on Additional Benchmarks}
\label{sec:mmsvg_bench}

To further evaluate the generalization capability of our method and reduce potential concerns regarding training data overlap, we additionally conduct experiments on \textbf{MMSVG-Bench}. This benchmark contains 300 GPT-generated prompts designed specifically for evaluating text-to-SVG generation models.

All methods are evaluated in a \textbf{zero-shot setting} without additional fine-tuning. As shown in Table~\ref{tab:mmsvg_bench}, IntroSVG achieves the best performance across all metrics, including CLIP-based text-image similarity (CLIP-T2I), aesthetic score, and HPSv1 preference score.

\begin{table}[h]
\vspace{-0.8em}
\centering
\caption{\small Zero-shot Results on MMSVG-Bench}
\label{tab:mmsvg_bench}
\resizebox{0.43\textwidth}{!}{
\begin{tabular}{lcccc}
\toprule
Method & Avg. Token & CLIP-T2I $\uparrow$ & Aes. $\uparrow$ & HPSv1 $\uparrow$ \\
\midrule
GPT-4o & 403.93 & 0.2301 & 4.6684 & 0.1835 \\
OmniSVG & 3698.87 & 0.1953 & 4.6554 & 0.1760 \\
SVGen & 1517.45 & 0.2235 & 4.5456 & 0.1879 \\
\textbf{IntroSVG (Ours)} & 1981.26 & \textbf{0.2456} & \textbf{4.8141} & \textbf{0.1901} \\
\bottomrule
\end{tabular}
}
\vspace{-1em}
\end{table}

\textbf{Unified Evaluation Set.}
For the main experiments in the paper, we construct a unified evaluation set containing \textbf{1,400 samples}. The dataset is stratified according to the training sources used by existing methods (LLM4SVG: 200 samples, OmniSVG: 600 samples, SVGen: 600 samples). All samples are strictly excluded from the training corpus to avoid any potential data overlap.

\subsection{Efficiency, Latency, and Cost Analysis}
\label{sec:efficiency}

We further analyze the computational efficiency of the proposed iterative generation framework. All experiments are conducted on a single NVIDIA H100 GPU using the \texttt{lmdeploy} inference engine.

Table~\ref{tab:rebuttal_efficiency} reports generation latency, token usage, and generation quality across different inference strategies.

\begin{table}[h]
\vspace{-0.8em}
\centering
\caption{\small Efficiency and Cost Analysis}
\label{tab:rebuttal_efficiency}
\resizebox{0.43\textwidth}{!}{
\begin{tabular}{lccccc}
\toprule
Method & FID $\downarrow$ & Aes. $\uparrow$ & Latency & Total Tokens \\
\midrule
Qwen3-VL-32B & 38.68 & 4.39 & 6.51 s & 445.75 \\
Iter 0 & 29.76 & 4.83 & 9.12 s & 1875.47 \\
Best-of-4 (Iter 0) & 28.43 & 4.85 & 39.01 s & 8522.64 \\
\textbf{IntroSVG (Iter 3)} & \textbf{26.18} & \textbf{4.89} & 42.31 s & 8840.29 \\
\bottomrule
\end{tabular}
}
\vspace{-1em}
\end{table}

The results demonstrate that iterative refinement significantly improves generation quality. Specifically, the FID score improves from 29.76 for the initial draft (Iter 0) to 26.18 after three refinement iterations.

We further compare our method with a \textbf{Best-of-4 sampling strategy}, which selects the best result from four independently generated candidates. Although Best-of-4 uses comparable computational resources, it still performs worse than our iterative refinement strategy. This result suggests that allocating computation to structured \emph{introspection and revision} is more effective than relying solely on stochastic sampling.

\subsection{Human Evaluation}
\label{sec:human_eval}

To further assess the perceptual quality of generated SVG graphics, we conduct a human evaluation study involving five professional designers.

We randomly sample \textbf{100 prompts} from the evaluation set and generate SVG results using different methods. All samples are evaluated in a \textbf{blind evaluation setting}, where annotators are not informed of the model identity.

\begin{figure}[tb]
\centering
\includegraphics[width=0.5\textwidth,height=0.19\textwidth]{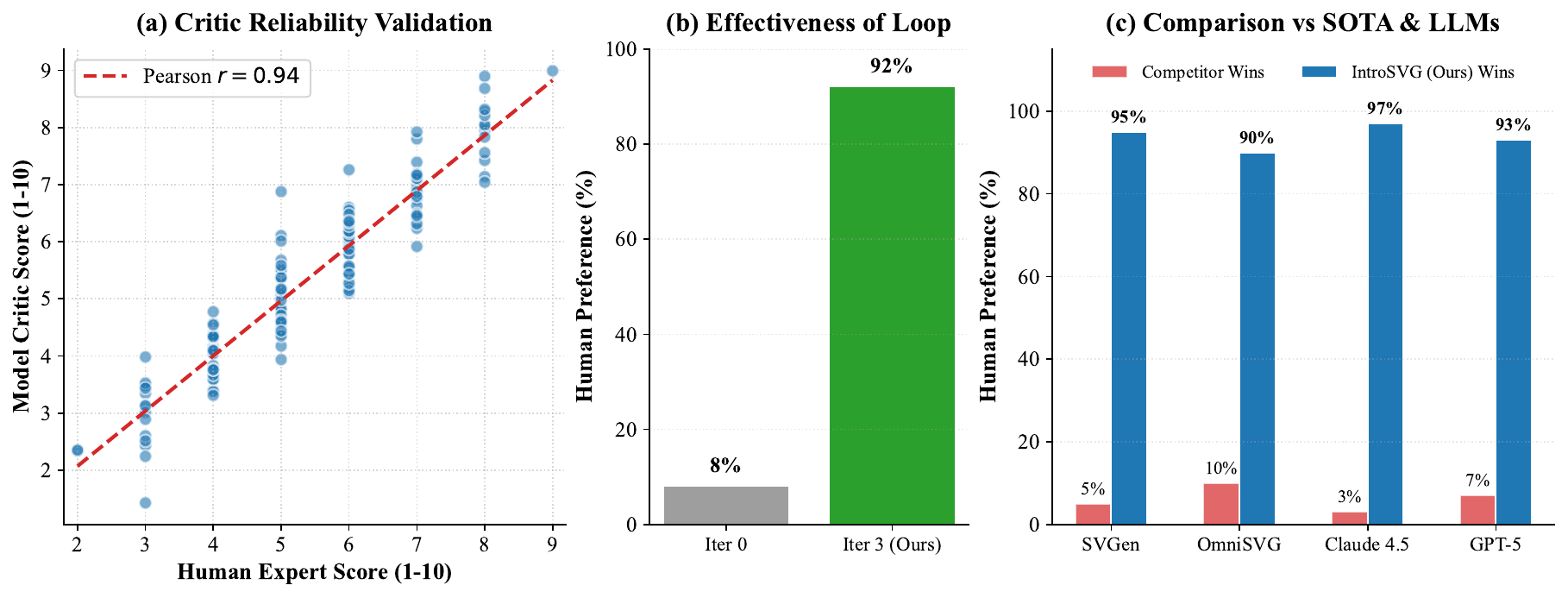}
\caption{Human evaluation results.}
\label{human}
\vspace{-2em}
\end{figure}

We conduct human evaluation using two protocols: (1) a 1–10 Likert scale to assess the reliability of Critic scores, and (2) pairwise blind A/B comparisons to evaluate the visual quality of model outputs. The results are analyzed as follows.

\textbf{Critic Reliability.}
Annotators rate each result using a \textbf{1–10 Likert scale} according to visual aesthetics and prompt alignment. The \textbf{Pearson correlation} between the Critic scores and human ratings reaches \textbf{0.94}, indicating strong agreement between the automated evaluation and human perception.

\textbf{Effectiveness of Iterative Refinement.}
As shown in Figure~\ref{human}(b), \textbf{92\%} of the samples demonstrate that Iteration 3 produces higher-quality results than the initial draft (Iter 0), highlighting the effectiveness of the refinement process.

\textbf{Comparison with Baseline Methods.}
In pairwise blind comparisons, IntroSVG achieves significantly higher win rates against existing methods, including SVGen (\textbf{95\%}), OmniSVG (\textbf{90\%}), GPT-5 (\textbf{93\%}), and Claude 4.5 (\textbf{97\%}). These results further confirm the superiority of the proposed iterative generation framework.

\section{Sample Demonstration}
\begin{figure*}[htbp]
    \centering
    \includegraphics[width=0.9\textwidth]{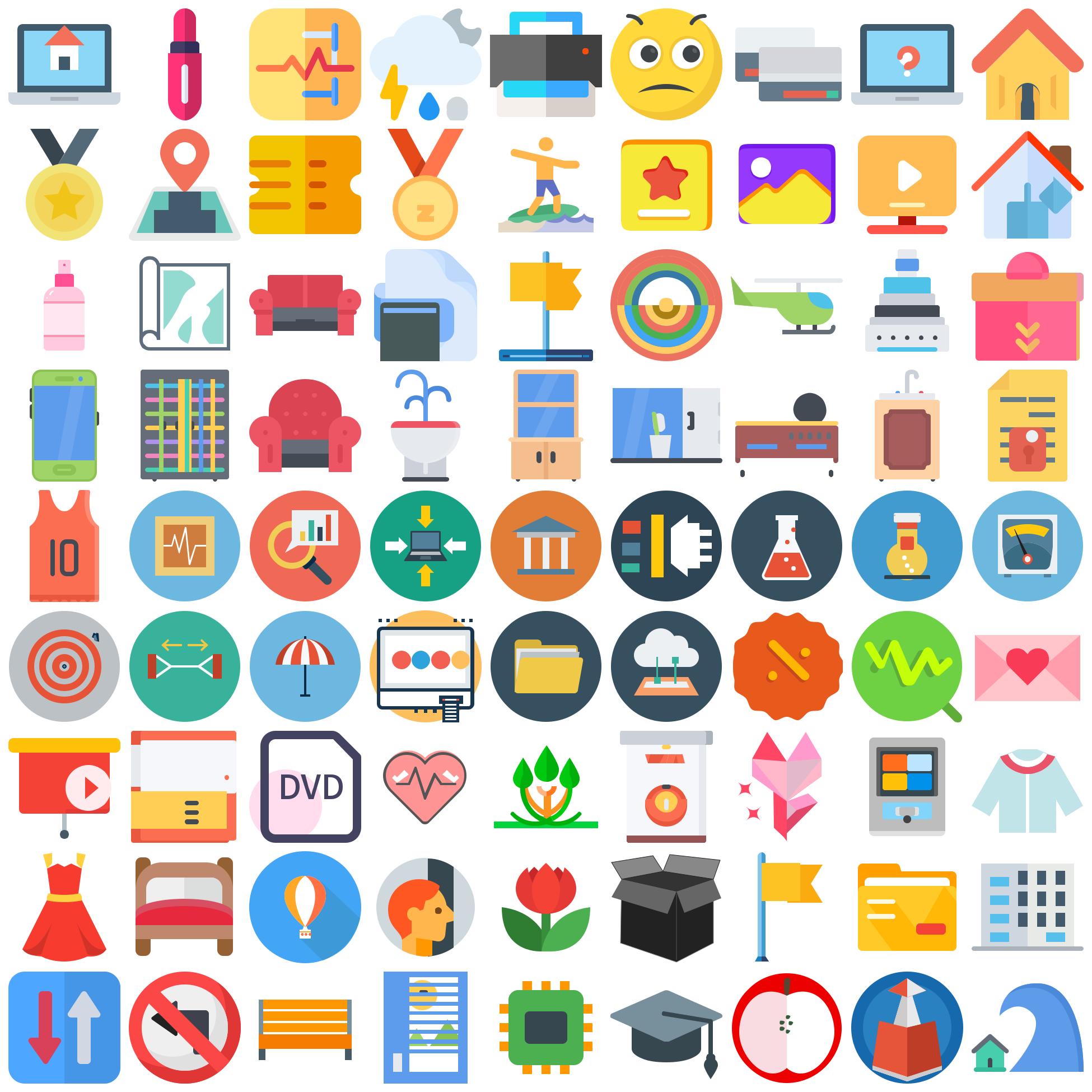}
    \caption{SVG samples generated by IntroSVG}
    \label{fig:gallery_res}
\end{figure*}
\subsection{Training Data Examples}
We first present samples of the synthetic critique data generated by GPT-4o in Figure~\ref{fig:critique_examples}, which serves as the ground truth for training the model's self-evaluation capability. Subsequently, Figure~\ref{fig:dataset_samples} illustrates the specific input-output formats constructed based on these data for the Generation, Correction, and Critique tasks.

\subsection{Generated SVG Results}
We display a collection of SVG icons generated by IntroSVG in Figure~\ref{fig:gallery_res}. The samples demonstrate the aesthetic performance of the model in generating complex, multicolored SVG icons.

\end{document}